\documentclass[runningheads]{llncs}

\usepackage{eccv}

\usepackage{eccvabbrv}

\usepackage{graphicx}
\usepackage{booktabs}
\usepackage{multirow}
\usepackage{xcolor}
\usepackage[accsupp]{axessibility}  

\usepackage{hyperref}

\usepackage{orcidlink}
\usepackage{marvosym}

\usepackage{footmisc} 

\begin{document}

\title{DAIT: Distillation from Vision-Language Models to Lightweight Classifiers with Adaptive Intermediate Teacher Transfer} 

\titlerunning{DAIT}

\author{Zhengxu He\inst{1} \and
Jun Li\inst{1} \href{\#fn:corresponding}{\textsuperscript{(\Letter)}}\and
Zhijian Wu\inst{2}}


\authorrunning{Z. He et al.}

\institute{School of Computer and Electronic Information, Nanjing Normal University, Nanjing, China \and
Medical Artificial Intelligence Lab, Westlake University}

\maketitle

\footnotetext[1]{
  \label{fn:corresponding}
  For correspondences please contact: \email{lijuncst@njnu.edu.cn}.
}

\begin{abstract}
  Large-scale Vision-Language Models (VLMs) encode rich multimodal semantics that are highly beneficial for fine-grained visual categorization (FGVC). However, their prohibitive computational cost hinders practical deployment in resource-constrained environments. Although knowledge distillation contributes to transferring VLMs capacity to lightweight classifiers, conventional distillation mechanisms, which directly transfer from a generic VLM to a compact student, often yield suboptimal results due to severe architectural misalignment and introducing task-irrelevant information. To alleviate this limitation, we propose Distillation with Adaptive Intermediate Teacher transfer (DAIT) in this study, facilitating adaptive knowledge transfer from VLMs to lightweight students. DAIT introduces a trainable intermediate teacher that learns to transfer frozen VLMs representations under explicit supervision from the target fine-grained task. This intermediate teacher adaptively enhances discriminative visual cues, thereby producing compact and task-aligned knowledge that can be reliably distilled into lightweight models. Extensive evaluations on multiple FGVC benchmarks with diverse student architectures demonstrate that our method achieves respective performance gains of 12.63\% and 8.34\% on FGVC-Aircraft and CUB-200-2011 datasets, establishing DAIT as a principled paradigm for transferring from general-purpose VLMS to deployable fine-grained recognition models.
  \keywords{Knowledge distillation \and Fine-grained visual categorization \and Vision-Language Models \and Adaptive intermediate teacher transfer }
\end{abstract}

\section{Introduction}
\label{sec:intro}

Fine-grained visual categorization (FGVC)\cite{berg2014birdsnap,maji2013fine} aims to distinguish highly similar subcategories, such as bird species\cite{2011The}. This task relies on subtle and localized discriminative cues\cite{xie2013hierarchical,huang2016part}, where inter-class differences are minimal and easily affected by background clutter and viewpoint variations. Consequently, FGVC places stringent demands on the representation capacity of recognition models. 

Large-scale vision–language models (VLMs)\cite{alayrac2022flamingo,cherti2023reproducible,li2022blip,radford2021learning,wang2023image,li2022grounded}, pretrained on massive image–text corpora, learn strong cross-modal semantics and fine-grained visual features, offering clear potential for FGVC. However, their substantial computational and memory costs hinder deployment on edge devices\cite{yang2024clip}, and retraining or deep fine-tuning per task is often impractical.

\begin{figure*}
\centering
\subfloat[]{
  \begin{minipage}[b]{0.23\linewidth}
    \centering 
    \includegraphics[width=1\linewidth]{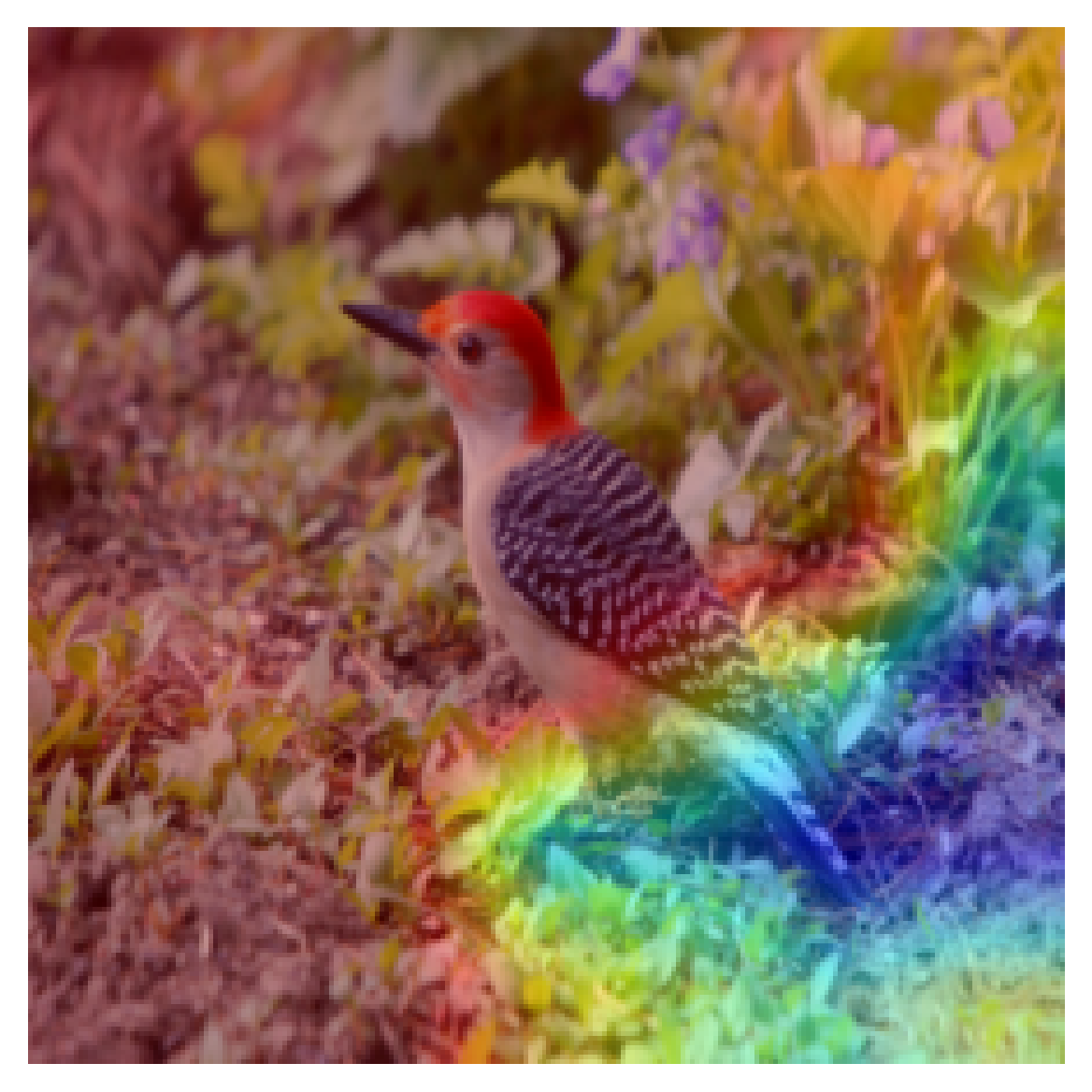}
    \includegraphics[width=1\linewidth]{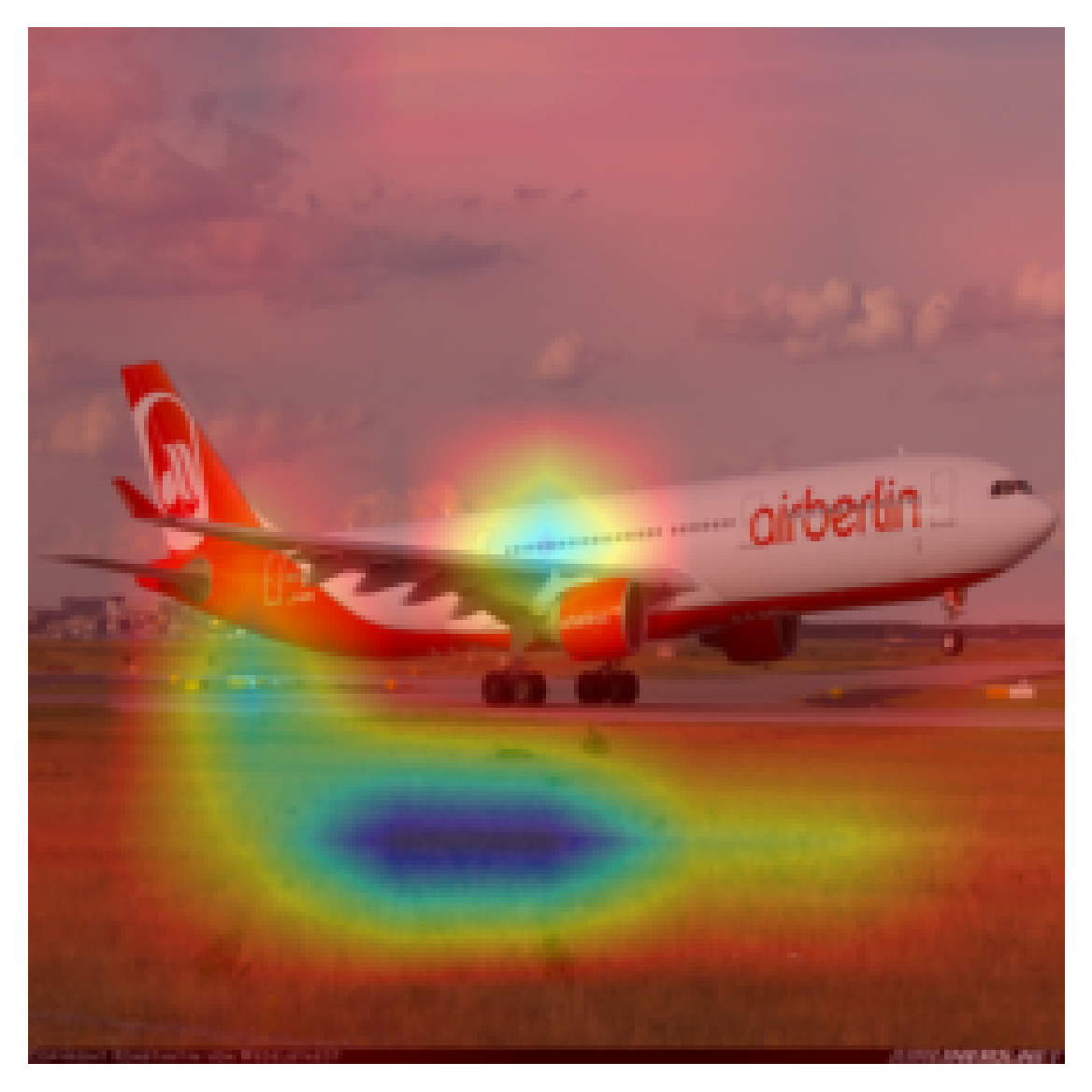}
  \end{minipage}
  \label{dis_clip}
}
\subfloat[]{
  \begin{minipage}[b]{0.23\linewidth}
    \centering
    \includegraphics[width=1\linewidth]{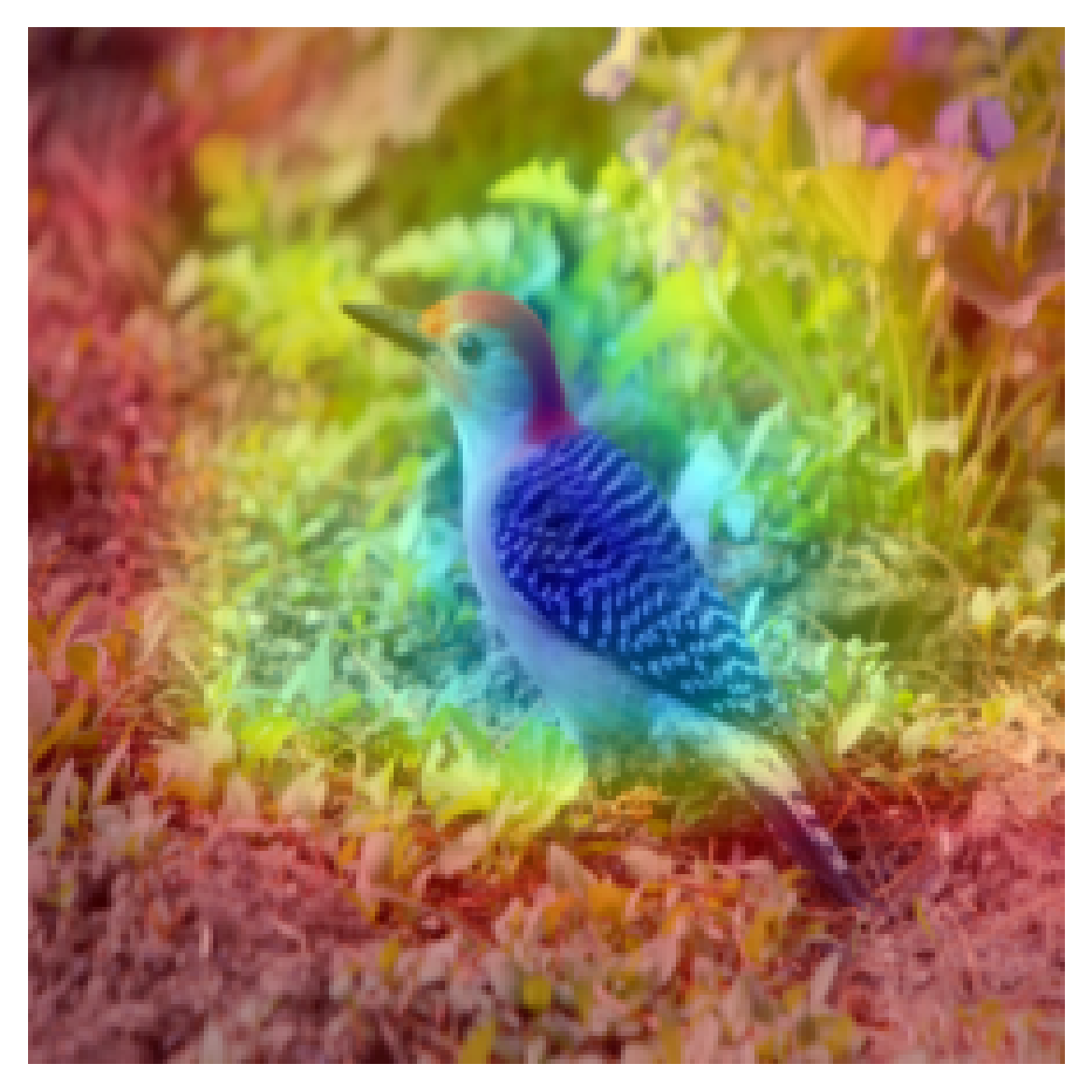}
    \includegraphics[width=1\linewidth]{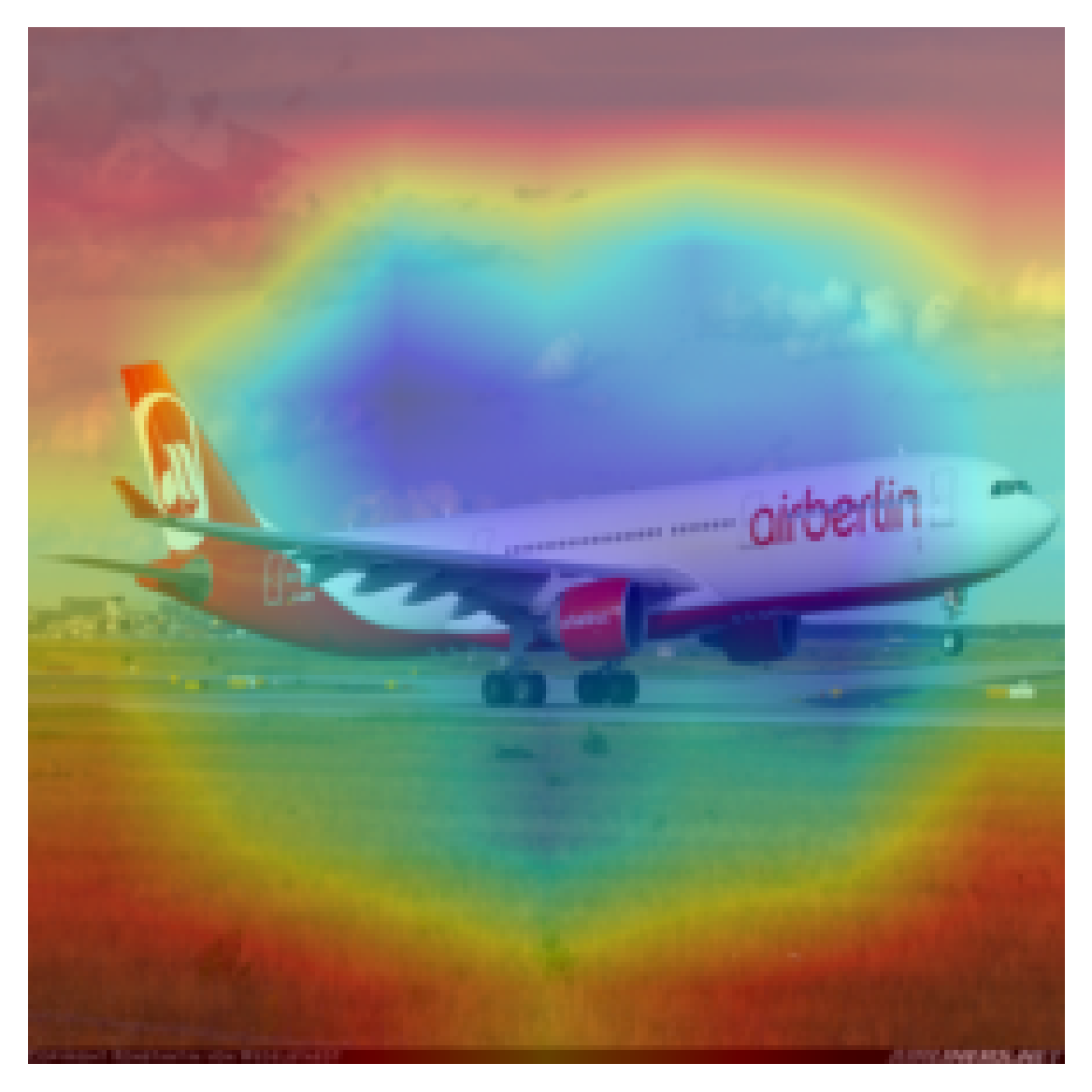}
  \end{minipage}
}
\subfloat[]{
  \begin{minipage}[b]{0.23\linewidth}
    \centering
    \includegraphics[width=1\linewidth]{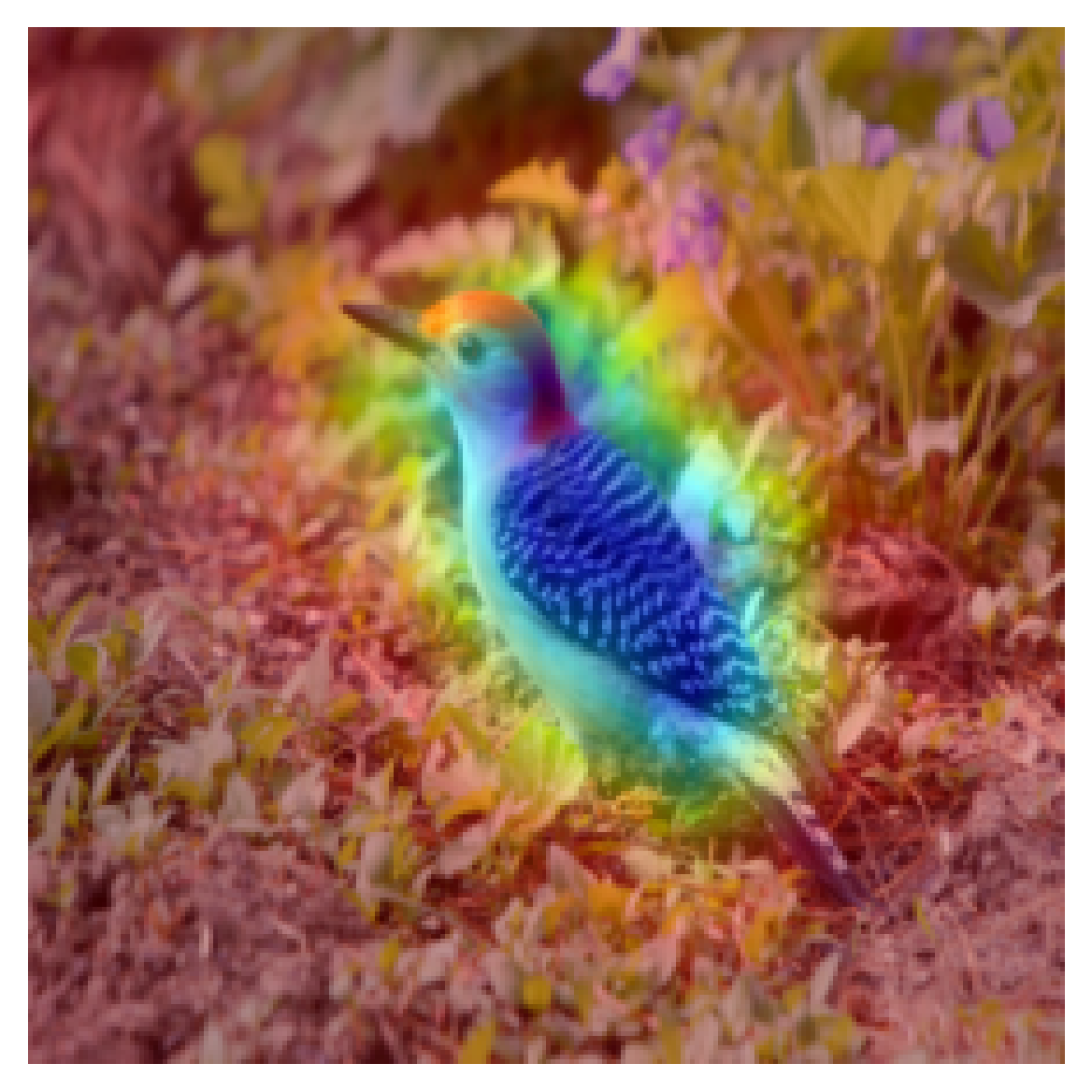}
    \includegraphics[width=1\linewidth]{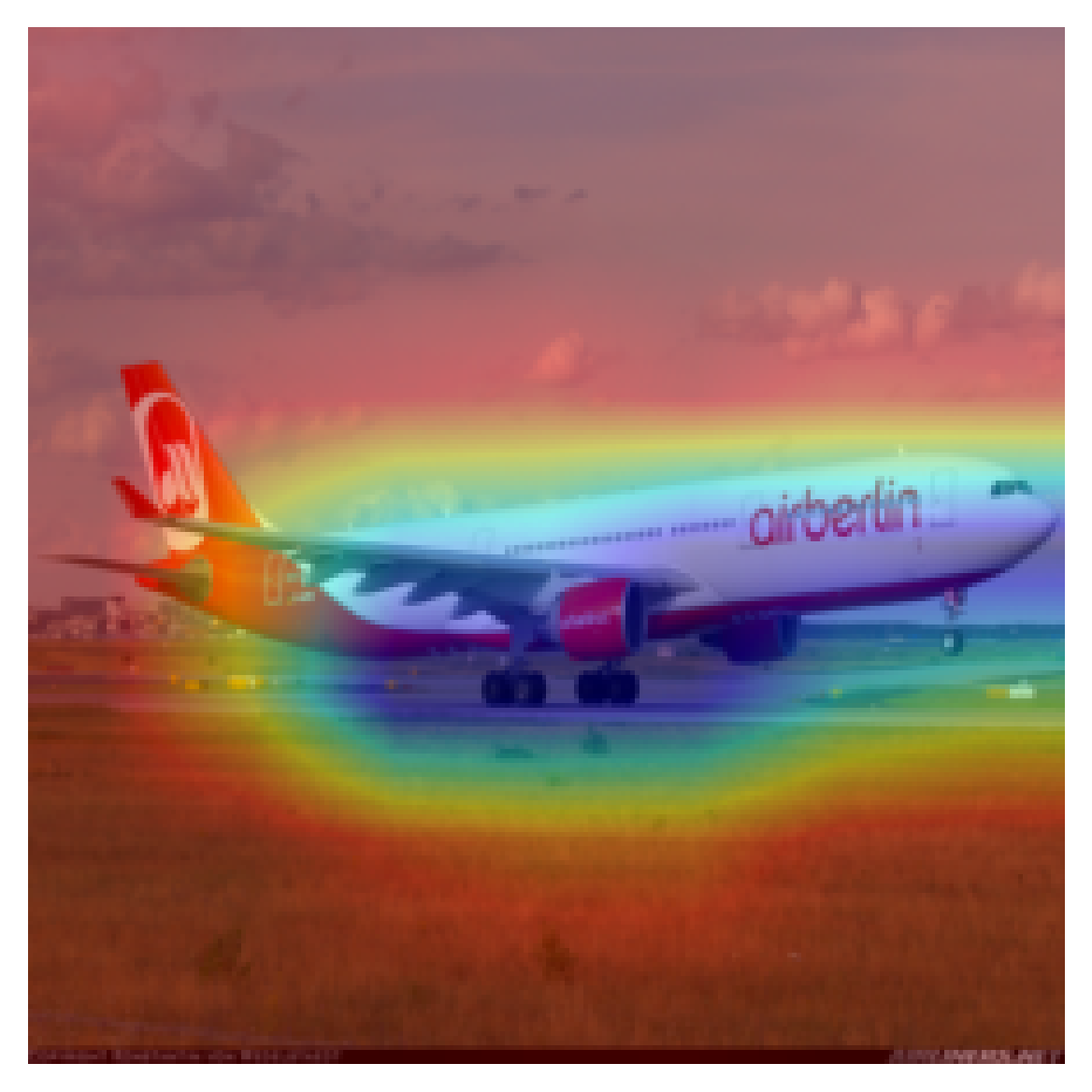}
  \end{minipage}
  \label{dis_reg}
}
\caption{Failure cases of direct distillation from VLM to lightweight student network ResNet18. (a) Direct distillation from the VLM. (b) Independently trained student without teacher guidance. (c) Our method: DAIT. Heatmaps are overlaid on the input images to indicate model attention. Compared to (a) and (b), our method yields more concentrated activations on discriminative parts (e.g., the bird's body) while suppressing background responses.}
\label{fig:pred_comparison}
\end{figure*}

Knowledge Distillation (KD)\cite{gou2021knowledge,heo2019knowledge} offers a natural solution to this challenge. By encouraging a student model to mimic the output\cite{hinton2015distilling}, intermediate features \cite{romero2014fitnets}, or relational structures\cite{park2019relational} of a teacher model, KD enables compact networks to inherit partial teacher’s representational ability while maintaining low computational complexity. KD has been widely adopted for model compression and cross-architecture transfer\cite{gou2021knowledge}. For FGVC, in particular, distillation is highly promising because the fine-grained semantic structures learned by powerful teachers provide richer supervision than hard labels alone.

However, when the teacher is a large-scale VLM and the student is a lightweight network, a substantial gap in representation capacity and modality arises\cite{fang2026knowledge}. It is usually difficult for small models to directly learn abundant high-dimensional and multimodal knowledge encoded in VLMs, often leading to unstable optimization or negative transfer in fine-grained tasks. Furthermore, VLMs are trained as general-purpose models and thus contain large amounts of task-irrelevant information. Directly distilling such unfiltered knowledge can introduce noisy or redundant supervision, weakening the student’s ability to focus on truly discriminative cues, as shown in \cref{dis_clip}.

To address these issues, we propose a novel Distillation framework with Adaptive Intermediate Teacher transfer termed DAIT. The core objective of DAIT is to enable efficient and task-adaptive knowledge transfer from large VLMs to lightweight students. DAIT introduces a trainable intermediate teacher between the VLM and the student, forming a task-adaptation layer. The intermediate teacher first learns the multimodal fine-grained representations from the VLM and is then optimized on the target task. As a result, its representations become more compact and more focused on task-relevant discriminative cues. The refined knowledge is subsequently distilled into the lightweight student model. The intermediate teacher not merely bridges the gap in model size, but rather functions as a knowledge filter and adapter. Without retraining VLM, it reorganizes the teacher’s knowledge in a task-oriented manner by suppressing irrelevant or noisy information and emphasizing fine-grained discriminative patterns. This hierarchical distillation strategy offers three advantages. First, it reduces an enormous gap in representational capacity between VLMs and students, leading to more stable training. Second, it provides more compact and task-specific discriminative supervision without introducing additional noise, improving learning efficiency and generalization. Third, it avoids frequent fine-tuning or retraining of large VLMs, substantially lowering practical deployment and maintenance costs. Extensive experiments on multiple standard fine-grained benchmarks using various lightweight student architectures. The results demonstrate that DAIT consistently outperforms conventional single-stage distillation methods in both accuracy and stability. To sum up, the contributions of this paper are threefold as follows:

\begin{itemize}
\item[$\bullet$] We propose DAIT, a distillation framework that enables efficient and adaptive task-specific knowledge transfer from large-scale VLMs to lightweight models for fine-grained visual categorization. 
\item[$\bullet$] Within our DAIT, the intermediate teacher achieves adaptive retraining and knowledge filtering, allowing multi-modal knowledge from VLMs to be reorganized into compact and discriminative supervision without requiring frequent retraining of the large models. 
\item[$\bullet$] We perform extensive experiments on multiple fine-grained datasets with diverse student architectures, demonstrating significant improvements in both accuracy and training stability over conventional distillation approaches.
\end{itemize}

\section{Related Work}

\subsection{Fine-Grained Visual Categorization}
Traditional FGVC methods seek more discriminative representations via part-based alignment\cite{zhang2014part}, attention\cite{fu2017look,vaswani2017attention}, multi-scale fusion\cite{lin2015bilinear}, and metric learning\cite{sun2018multi}, whereas they usually rely on deep computationally intensive backbones that limit practical deployment on edge devices. Consequently, recent work has explored efficiency-oriented architectures and micro-level optimizations—such as lightweight convolutional operators and structured pruning\cite{xia2022structured,mehta2021mobilevit}—to retain discriminative capacity under strict resource budgets. To further boost compact models, knowledge distillation has been widely used in FGVC, typically via feature imitation, attention transfer\cite{zagoruyko2016paying}, or relational constraints to provide richer supervision than labels alone. However, when teachers are large, generic, or multimodal, their representations may carry substantial task-irrelevant content, which is detrimental to knowledge transfer during the distillation process. This is precisely the motivation behind our proposal of adaptive intermediate teacher transfer.

\subsection{Knowledge Distillation}
Knowledge distillation\cite{cho2019efficacy,stanton2021does} was originally formulated as encouraging students to mimic the softened output distributions of teacher models. Since the pioneering work\cite{hinton2015distilling}, it has been extended to feature-level distillation\cite{romero2014fitnets}, attention transfer, and relational distillation\cite{park2019relational}. Feature-based methods constrain intermediate representations to preserve the discriminative structure of the teacher, while relational approaches further enforce consistency among samples or channels. These paradigms have been widely adopted for model compression and cross-architecture knowledge transfer.

With the rise of large-scale pretrained and multimodal models, considerable effort has been devoted to transferring their powerful capability to lightweight networks\cite{ma2023borrowing,huang2023sentence,dong2025llm}. Methods such as VL2Lite\cite{jang2025vl2lite} directly distill features from frozen VLMs into compact classifiers, by leveraging abundant multimodal semantics. However, it inevitably introduces task-irrelevant information that may impair fine-grained recognition performance. In contrast, our DAIT maintains frozen VLMs while introduces a trainable intermediate teacher that adaptively transfers VLM knowledge under fine-grained supervision. By reshaping and filtering VLM representations to emphasize local discriminative patterns and suppress irrelevant semantics, the intermediate teacher produces compact, task-aligned embeddings that enable more stable and effective distillation for lightweight students.

\section{Method}
\begin{figure}[!t]
    \centering
    \includegraphics[width=1\linewidth]{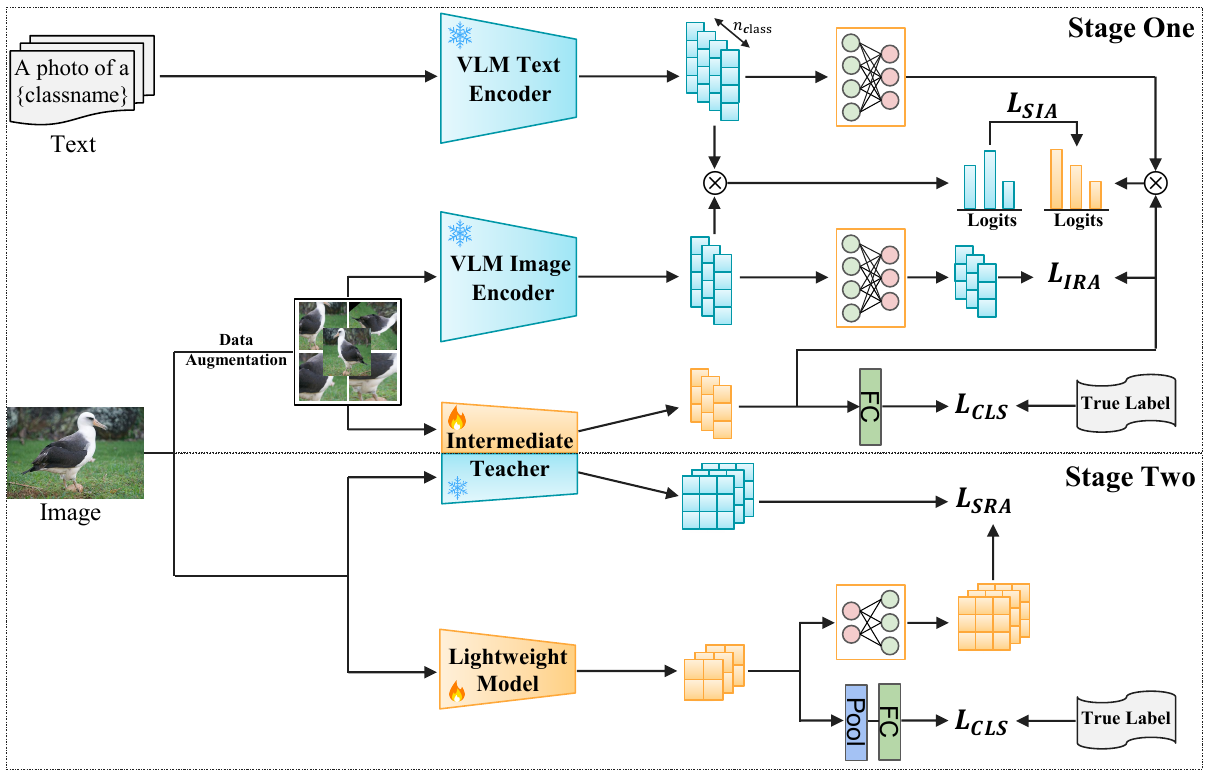}
    \caption{Overview of our proposed DAIT for distilling from a frozen VLM to a lightweight model via a trainable intermediate teacher model. With the help of data augmentation, the intermediate teacher captures rich fine-grained cues from the VLM outputs and performs adaptive knowledge transfer. It then processes original image features to produce refined and task-aligned supervision, which is transferred to the lightweight student by feature-level distillation.}
    \label{fig:frame}
\end{figure}

In this study, we propose a novel distillation framework termed DAIT. To bridge the tremendous gap between VLMs and lightweight models, we introduce an intermediate teacher to facilitate adaptive knowledge transfer. The overall architecture of the proposed DAIT is shown in \cref{fig:frame}.

\subsection{Distillation from VLM to Intermediate Teacher}
Given the strong visual representation capability of VLMs, we first apply data augmentation to the original input images $x$ to generate diverse training samples:
\begin{equation}
\tilde{x} = A(x),
\end{equation}
where $A(\cdot)$ is the data augmentation function and $\tilde{x}$ denotes the augmented samples of the original input images.

Subsequently, the augmented image samples $\tilde{x}$ are fed into the visual encoder of the VLM and the intermediate teacher model, respectively. Meanwhile, the prompt text corresponding to class $c$, denoted as $\mathrm{prompt}(c)$, passes through the text encoder of the VLM yielding auxiliary semantic embeddings. Considering the high dimensionality of the VLM output features, we perform a non-linear mapping to project the VLM features into a lower-dimensional space, thereby reducing the computational overhead of the model:
\begin{equation}
z_v = f_{vlm}(E_V^I(\tilde{x})),
\end{equation}

\begin{equation}
t_c = f_{vlm}(E_V^T(\mathrm{prompt}(c))),
\end{equation}
where $z_v$ and $t_c$ denote the outputs of the image encoder and the text encoder of the VLM after the condensation projection, respectively. $f_{vlm}(\cdot)$ denotes a two-layer multilayer perceptron, $z_v \in \mathbb{R}^{B \times D}$, $t_c \in \mathbb{R}^{N \times D}$, $B$ represents the batch size, $N$ is class number, and $D$ is feature dimension.
\begin{equation}
\tilde{z}_t = T_m(\tilde{x}),
\end{equation}
where $\tilde{z}_t$ represents the adaptively transferred feature vector of the VLM, $\tilde{z}_t \in \mathbb{R}^{B \times D}$ and $T_m(\cdot)$ denotes the image encoder of the intermediate teacher model.

In terms of training loss, two terms are involved, namely Semantic Image Alignment Loss (SIA) and Image Representation Alignment Loss (IRA).

\subsubsection{Semantic Image Alignment Loss.}The first term employs the Kullback–Leibler (KL) divergence to constrain the features of the intermediate teacher. We compute the cosine similarities between the image features and the textual features:
\begin{equation}
cos(z_v,t_c) = \frac{z_v t_c^\top}{\|z_{v}\| \|t_{c}\|}, 
\end{equation}
\begin{equation}
cos(\tilde{z}_t,t_c) = \frac{\tilde{z}_t t_c^\top}{\|\tilde{z}_{t}\| \|t_{c}\|}, 
\end{equation}
The loss is formulated as:
\begin{equation}
\mathcal{L}_{\text{SIA}} = T^2 \cdot \text{KL}\left(\text{softmax}\left(\frac{\cos{(\tilde{z}_t,t_c)}}{T}\right), \text{softmax}\left(\frac{\cos{(z_v,t_c)}}{T}\right)\right),
\end{equation}
where  $T$  is a temperature parameter that softens the probability distributions, facilitating the distillation of complex linguistic patterns into the intermediate teacher model.

\subsubsection{Image Representation Alignment Loss.}The second term adopts the L1 loss to further calibrate the representation of the intermediate teacher with the visual encoding of VLM, encouraging the mid-stage model to inherit strong representational capability from the VLM. Mathematically, the loss is formulated as:
\begin{equation}
\mathcal{L}_{\text{IRA}} = \frac{1}{BD} \sum_{i=1}^{B} \sum_{j=1}^{D} \left| \tilde{z}_t(i,j)- z_v(i,j) \right|,
\end{equation}

\subsubsection{Classification Loss.}In addition to $\mathcal{L}_{SIA}$ and $\mathcal{L}_{IRA}$, additional classification loss is introduced to regularize adaptive knowledge transfer from VLM, thereby encouraging the intermediate model to learn more discriminative and fine-grained visual representations. The classification loss is mathematically defined as:
\begin{equation}
\mathcal{L}_{\text{CLS}} = -\sum_{i=1}^{N} y_{i} \log(p(y_{i} \mid \tilde{x})), \label{eq9}  
\end{equation}
where $y_i$ is a binary indicator for the ground-truth class $i$, $p(y_i | \tilde{x})$ is the predicted probability of the class $i$ given input $\tilde{x}$.

\subsubsection{Overall Loss.}For training the intermediate teacher, the total loss $\mathcal{L}_{\text{VLM2Inter}}$ is defined as:\label{3.1}

\begin{equation}
\mathcal{L}_{\text{VLM2Inter}} = \lambda \mathcal{L}_{CLS} + \frac{(1-\lambda)}{2}(\mathcal{L}_{SIA} + \mathcal{L}_{IRA}),
\end{equation}
where $\lambda$ is a hyperparameter controlling the relative importance of
each loss component. Division by two ensures the total weight of all loss terms sums to 1.

In our experiments, $\lambda$ is initially set as 0. During training, $\lambda$ gradually increases, amplifying the importance of the classification loss. Its mathematically defined as:
\begin{equation}
    \lambda = ke + b,
\label{eq11}
\end{equation}

where $e$ denotes the current training epoch, $k$ is the linear rate of change, and 
$b$ denotes the bias term.

This dynamic weighting strategy ensures an initial focus on distilling knowledge from the VLM, followed by a refined emphasis on adaptive knowledge transfer for fine-grained recognition tasks as the model evolves.

\subsection{Distillation from Intermediate Teacher to Lightweight Model}After completing the adaptive knowledge transfer in the previous stage, the intermediate teacher is frozen and employed to guide the lightweight student model in the subsequent distillation process. Specifically, the original image samples are delivered to the intermediate teacher and student networks, leading to respective feature maps. Moreover, we impose a convolutional layer on the student feature for the subsequent alignment as follows:
\begin{equation}
z_s = f_{stu}(LW(x)),
\end{equation}
where $LW(x)$ is the feature encoding of the lightweight model, $f_{stu}$ denotes a convolutional layer, and $z_s \in \mathbb{R}^{B \times D \times H \times W}$.

\subsubsection{Spatial Representation Alignment Loss.}To facilitate the feature alignment, the feature maps of the final convolutional layer are adopted to bridge the teacher and the student using MSE loss, such that the student can learn fine-grained spatial cues from the intermediate model, which is mathematically formulated as:
\begin{equation}
\mathcal{L}_{\text{SRA}} = \frac{1}{HW}  \sum_{h=1}^{H} \sum_{w=1}^{W} \left\| z_s(h,w) - z_t(h,w) \right\|_2^2,
\end{equation}
where $z_t$ denotes the intermediate teacher feature map.

\subsubsection{Classification Loss.}Consistent with the loss of discriminative learning in the previous distillation stage, this loss encourages the lightweight model to produce high probabilities for the correct classes, improving its accuracy on the fine-grained visual classification task. Its mathematical formulation is formulated as \cref{eq9} .

\subsubsection{Overall Loss.}The total loss for the training phase when transferring from intermediate model to the lightweight student is defined as:
\begin{equation}
\mathcal{L}_{\text{Inter2Lite}} = \lambda \mathcal{L}_{CLS} + (1- \lambda)\mathcal{L}_{SRA},
\end{equation}
where $\lambda$ is consistent with that in the overall loss defined in  \cref{3.1}.

\subsection{Comparison with Prior Works}
Although several previous studies also attempt to improve the distillation performance via multi-stage distillation, our method is fundamentally different from them in the following aspects. 

First, prior works such as TAKD\cite{mirzadeh2020improved}, DGKD\cite{son2021densely}, AMD\cite{han2024amd} mainly operate within homogeneous architectures and reduce the teacher-student gap in representational capability by only introducing intermediate models with progressively reduced size (e.g., a family of CNNs of decreasing depth: 10→8→6→4→2 layers). In contrast, our DAIT is targeted at distillation from large-scale pretrained foundation models, particularly VLMs, to compact student classifiers for FGVC. In this setting, the teacher-student disparity is not limited to model size but also manifested in architectural misalignment, representation discrepancy, and distribution shift, which poses significant challenges to knowledge transfer. Since VLMs encode generic multimodal semantics that contain substantial task-irrelevant information, directly transferring from VLMs to lightweight CNN models leads to suboptimal results in previous research such as VL2Lite. 

Second, our DAIT formulates the intermediate teacher as a task adapter and knowledge filter that transfers frozen VLM representations into compact and task-aligned features under fine-grained supervision. This more emphasizes knowledge transfer rather than teacher cascading, allowing the student to learn discriminative representations instead of directly imitating raw foundation-model outputs. This is in contrast to previous work in which the intermediate model only serves as a structural assistant for consecutive compression.

Third, while earlier works (e.g., TAKD) are mainly validated through logit-level distillation in homogeneous CNN setting, we evaluate both feature-level and output-level transfer in the cross-paradigm scenario of VLM-to-lightweight distillation. By maintaining frozen VLM and introducing a lightweight trainable intermediate teacher, our DAIT avoids repeated fine-tuning of expensive models and supports efficient deployment for real-world FGVC applications.

\section{Experiments}
\subsubsection{Datasets.}For evaluation, a wide range of public benchmarking datasets are involved for fine-grained visual classification. These datasets include CUB-200-2011\cite{2011The} for bird species categorization, FGVC-Aircraft\cite{maji2013fine} for identifying aircraft models, and NABirds\cite{van2015building} for species recognition. In addition, the Stanford Cars dataset\cite{Sfcars20133d} requires detailed distinctions among car models, whilst the Stanford Dogs dataset\cite{khosla2011novel} provides a basis for breed identification in dogs. We adhered to established dataset splits to maintain consistency with conventional studies.
\subsubsection{Implementation Details.}For the Vision-Language Model, we employ the
ConvNeXt-XXLarge network\cite{liu2022convnet} from OpenCLIP\cite{cherti2023reproducible}, which generates a 1024-dimensional feature output. For the intermediate teacher, we select the RegNet-Y-1.6GF network\cite{radosavovic2020designing} based on comparative evaluations, with detailed analyzes provided in the  \cref{inter_teacher}. Moreover, to assess the effectiveness of our method, we employed multiple lightweight architectures such as ResNet-18 (R-18)\cite{he2016deep}, EfficientNet-B0 (Eff-B0)\cite{tan2019rethinking}, MobileNet-V2 (Mob-V2)\cite{sandler2018mobilenetv2} and ShuffleNet-V2 (Shuf-V2)\cite{ma2018shufflenet}. These models were initialized with ImageNet-pretrained weights. 
For performance measure, the top-1 accuracy is utilized to assess the effectiveness of the proposed method. 
In implementation, all models for 100 epochs with AdamW optimizer, where the weight decay is 0.0001. The learning rate is initialized to 0.0001 and decayed for every 30 epochs. The batch size is set to 32. The temperature parameter $T$ is set to 2. The setting of the hyperparameter $\lambda$ is detailed in \cref{hyper}. Each image is resized to $224\times224$. All experiments are conducted on machines equipped with a single NVIDIA 3090 GPU.

\begin{table}[!b]
\centering  
\caption{Results of fine-grained classification performance on various datasets using ResNet-18 and MobileNet-V2 with different knowledge distillation methods. “w/o KD” denotes training without teacher guidance and is used to verify any performance difference attributed to the knowledge distillation process.}
\label{tab:1}

\resizebox{\columnwidth}{!}{
\begin{tabular}{l c ccccc}
\toprule
\multirow{2}{*}{Model} & \multirow{2}{*}{Method} & \multicolumn{5}{c}{Datasets} \\

& & CUB-200 & Aircraft & Sf Dogs & Sf Cars & NABirds\\
\midrule
\multirow{9}{*}{R-18}
& w/o KD & 64.95 & 50.98 & 67.80 & 70.03 & 57.01 \\
& KD\cite{hinton2015distilling} (T: R-152) & 65.43 & 51.94 & 67.37 & 72.07 & 57.05 \\
& RKD\cite{park2019relational} (T: R-152) & 66.49 & 54.58 & 68.80 & 72.96 & 58.87  \\
& KD\cite{hinton2015distilling} (T: CLIP) & 70.95 & 53.83 & 68.80 & 74.99 & 62.22 \\
& RKD\cite{park2019relational} (T: CLIP) & 68.31 & 50.98 & 69.03 & 72.19 & 61.16 \\
& RISE\cite{huang2023sentence} & 69.69 & 54.81 & 68.72 & 72.45 & 59.00  \\
& BorLan\cite{ma2023borrowing} & 66.79 & 51.14 & 69.25 & 70.71 & 59.94  \\
& VL2Lite\cite{jang2025vl2lite} & 71.38 & 55.87 & 72.40 & 77.09 & 63.26  \\
& \textbf{DAIT (Ours)} & \textbf{79.77} & \textbf{67.44} & \textbf{78.10} & \textbf{88.96} & \textbf{74.38}  \\
\midrule
\multirow{9}{*}{Mob-V2}
& w/o KD & 64.13 & 48.54 & 68.14 & 68.52 & 57.79 \\
& KD\cite{hinton2015distilling} (T: R-152) & 63.48 & 49.52 & 69.02 & 69.31 & 57.41 \\
& RKD\cite{park2019relational} (T: R-152) & 66.09 & 52.39 & 69.55 & 70.89 & 59.23 \\
& KD\cite{hinton2015distilling} (T: CLIP) & 68.00 & 48.78 & 73.02 & 68.74 & 63.81\\
& RKD\cite{park2019relational} (T: CLIP) & 67.95 & 50.29 & 69.45 & 71.23 & 60.93 \\
& RISE\cite{huang2023sentence} & 67.51 & 52.44 & 69.28 & 72.67 & 60.43 \\
& BorLan\cite{ma2023borrowing} & 67.38 & 52.69 & 70.62 & 69.08 & 62.16 \\ 
& VL2Lite\cite{jang2025vl2lite} & 71.02 & 53.82 & 73.24 & 74.99 & 64.21 \\ 
& \textbf{DAIT (Ours)} & \textbf{79.52} & \textbf{67.47} & \textbf{78.64} & \textbf{88.59} & \textbf{74.34} \\
\bottomrule
\end{tabular}
}
\end{table}

\begin{table}[!t]
\centering
\caption{Comparative results between DAIT and the suboptimal method VL2Lite across multiple datasets and different models. DAIT-F and DAIT-L denote our DAIT employing feature and logits distillation mechanisms, respectively.}
\label{tab:2}

\resizebox{\columnwidth}{!}{
\begin{tabular}{l c ccccc}
\toprule
\multirow{2}{*}{Model} & \multirow{2}{*}{Method} & \multicolumn{5}{c}{Datasets} \\

& &  CUB-200 & Aircraft & Sf Dogs & Sf Cars & NABirds \\
\midrule
\multirow{3}{*}{R-18}
& VL2Lite & 71.38 & 55.87 & 72.40 & 77.09 & 63.26 \\
& \textbf{ DAIT-F } & 79.77 \scriptsize{\textcolor{blue}{+8.39}} & 67.44 \scriptsize{\textcolor{blue}{+11.57}} & 78.10 \scriptsize{\textcolor{blue}{+5.70}} & 88.96 \scriptsize{\textcolor{blue}{+11.87}} & 74.38 \scriptsize{\textcolor{blue}{+11.12}} \\
& \textbf{ DAIT-L } & 76.46 \scriptsize{\textcolor{red}{+5.08}} & 65.08 \scriptsize{\textcolor{red}{+ 9.21}} & 75.78 \scriptsize{\textcolor{red}{+3.38}} & 86.84 \scriptsize{\textcolor{red}{+ 9.75}} & 71.14 \scriptsize{\textcolor{red}{+ 7.88}} \\
\midrule
\multirow{3}{*}{Mob-V2}
& VL2Lite & 71.02 & 53.82 & 73.24 & 74.99 & 64.21 \\
& \textbf{ DAIT-F } & 79.52 \scriptsize{\textcolor{blue}{+8.50}} & 67.47 \scriptsize{\textcolor{blue}{+13.65}} & 78.64 \scriptsize{\textcolor{blue}{+5.40}} & 88.59 \scriptsize{\textcolor{blue}{+13.60}} & 74.34 \scriptsize{\textcolor{blue}{+10.13}} \\
& \textbf{ DAIT-L } & 77.72 \scriptsize{\textcolor{red}{+6.70}} & 67.21 \scriptsize{\textcolor{red}{+13.39}} & 77.62 \scriptsize{\textcolor{red}{+4.38}} & 87.00 \scriptsize{\textcolor{red}{+12.01}} & 72.50 \scriptsize{\textcolor{red}{+ 8.29}} \\
\midrule
\multirow{3}{*}{Shuf-V2}
& VL2Lite & 74.99 & 54.88 & 76.59 & 78.11 & 69.78 \\
& \textbf{ DAIT-F } & 82.56 \scriptsize{\textcolor{blue}{+7.57}} & 69.25 \scriptsize{\textcolor{blue}{+14.37}} & 80.69 \scriptsize{\textcolor{blue}{+4.10}} & 90.02 \scriptsize{\textcolor{blue}{+11.91}} & 76.47 \scriptsize{\textcolor{blue}{+6.69}} \\
& \textbf{ DAIT-L } & 81.32 \scriptsize{\textcolor{red}{+6.33}} & 67.33 \scriptsize{\textcolor{red}{+12.45}} & 80.09 \scriptsize{\textcolor{red}{+3.50}} & 88.94 \scriptsize{\textcolor{red}{+10.83}} & 75.63 \scriptsize{\textcolor{red}{+5.85}} \\
\midrule
\multirow{3}{*}{Eff-B0}
& VL2Lite & 73.87 & 52.18 & 77.68 & 75.00 & 69.78 \\
&\textbf{ DAIT-F } & 82.76 \scriptsize{\textcolor{blue}{+8.89}} & 70.99 \scriptsize{\textcolor{blue}{+18.81}} & 81.96 \scriptsize{\textcolor{blue}{+4.28}} & 90.33 \scriptsize{\textcolor{blue}{+15.33}} & 78.42 \scriptsize{\textcolor{blue}{+8.64}} \\
& \textbf{ DAIT-L } & 82.00 \scriptsize{\textcolor{red}{+8.13}} & 70.63 \scriptsize{\textcolor{red}{+18.45}} & 81.38 \scriptsize{\textcolor{red}{+3.70}} & 89.07 \scriptsize{\textcolor{red}{+14.07}} & 77.49 \scriptsize{\textcolor{red}{+7.71}} \\
\bottomrule
\end{tabular}
}
\end{table}

\subsubsection{Main Results.}The results presented in \cref{tab:1} and  \cref{tab:2} show the effectiveness of DAIT in enhancing the performance of lightweight models through adaptive intermediate teacher transfer. By comparing DAIT with traditional KD methods using teacher model as ResNet152\cite{he2016deep} and CLIP\cite{radford2021learning}, it is evident that DAIT achieves superior performance, underscoring its efficacy in distilling complex multimodal knowledge into more compact models.

\begin{figure*}[!t]
    \centering
    \includegraphics[width=0.8\textwidth]{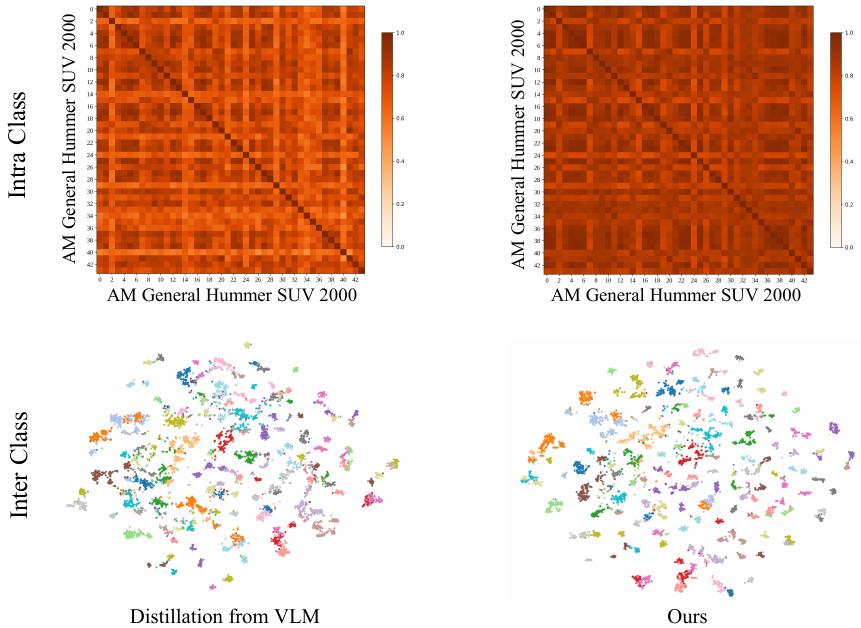}
    \caption{Visualization of intra-class and inter-class feature distributions of ResNet-18 distilled by different methods on the Stanford Cars dataset. They are illustrated via similarity matrices and t-SNE plots. With the guidance of an intermediate teacher, our method enables ResNet-18 to attain stronger discriminative capability.}
    \label{fig:fig3}
\end{figure*}

\subsubsection{Comparison of Visualized Feature Distributions.}To better illustrate the effect of DAIT on fine-grained visual classification, we visualize the feature distributions of ResNet-18 on the Stanford Cars dataset in terms of intra-class compactness and inter-class separability using relation matrices and t-SNE\cite{van2008visualizing}, as shown in \cref{fig:fig3}. The results demonstrate that DAIT simultaneously increases inter-class distances and promotes tighter intra-class clustering, yielding more discriminative feature representations.

\section{Ablation Study}
\subsubsection{The Choice of Intermediate Teacher.}\label{inter_teacher}
\begin{figure*}[!t]
    \centering
    \includegraphics[width=0.9\textwidth]{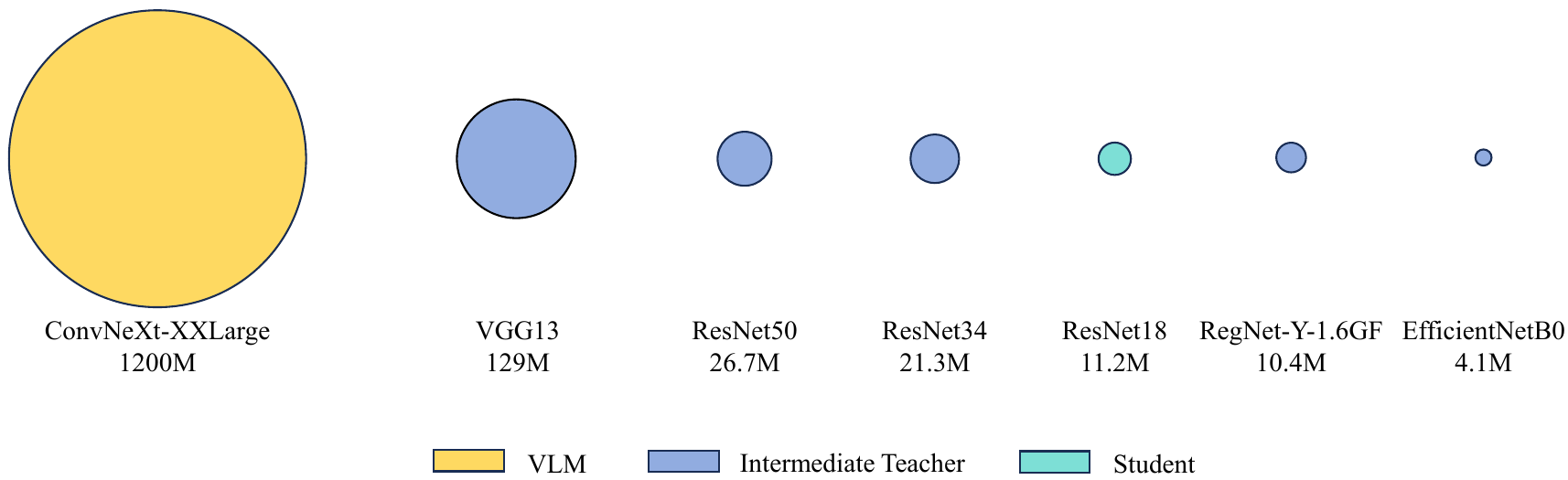}
    \caption{Bubble visualization of parameter counts for different deep architectures. The number below each bubble denotes the model size in millions (M). Unlike prior works that usually use a network larger than the student as a teacher assistant, our DAIT adopts even smaller than the student model as the intermediate teacher.}
    \label{fig:fig5}
\end{figure*}

We select several representative off-the-shelf network architectures as intermediate teachers for comparison, including ResNet-50\cite{he2016deep}, ResNet-34\cite{he2016deep}, EfficientNet-B0, RegNet-Y-1-6GF, and VGG-13\cite{simonyan2014very}. The student model is ResNet-18, while the large teacher model is ConvNeXt-XXLarge, as shown in \cref{fig:fig5}.
These candidate models cover a wide range of architectural paradigms, including traditional deep residual networks (ResNet), classical stacked networks (VGG), lightweight designs (EfficientNet), and scalable convolutional families (RegNet). This diversity enables a comprehensive analysis of the impact of intermediate teachers from an architectural perspective.

\begin{figure*}[!b]
    \centering
    \includegraphics[width=1\textwidth]{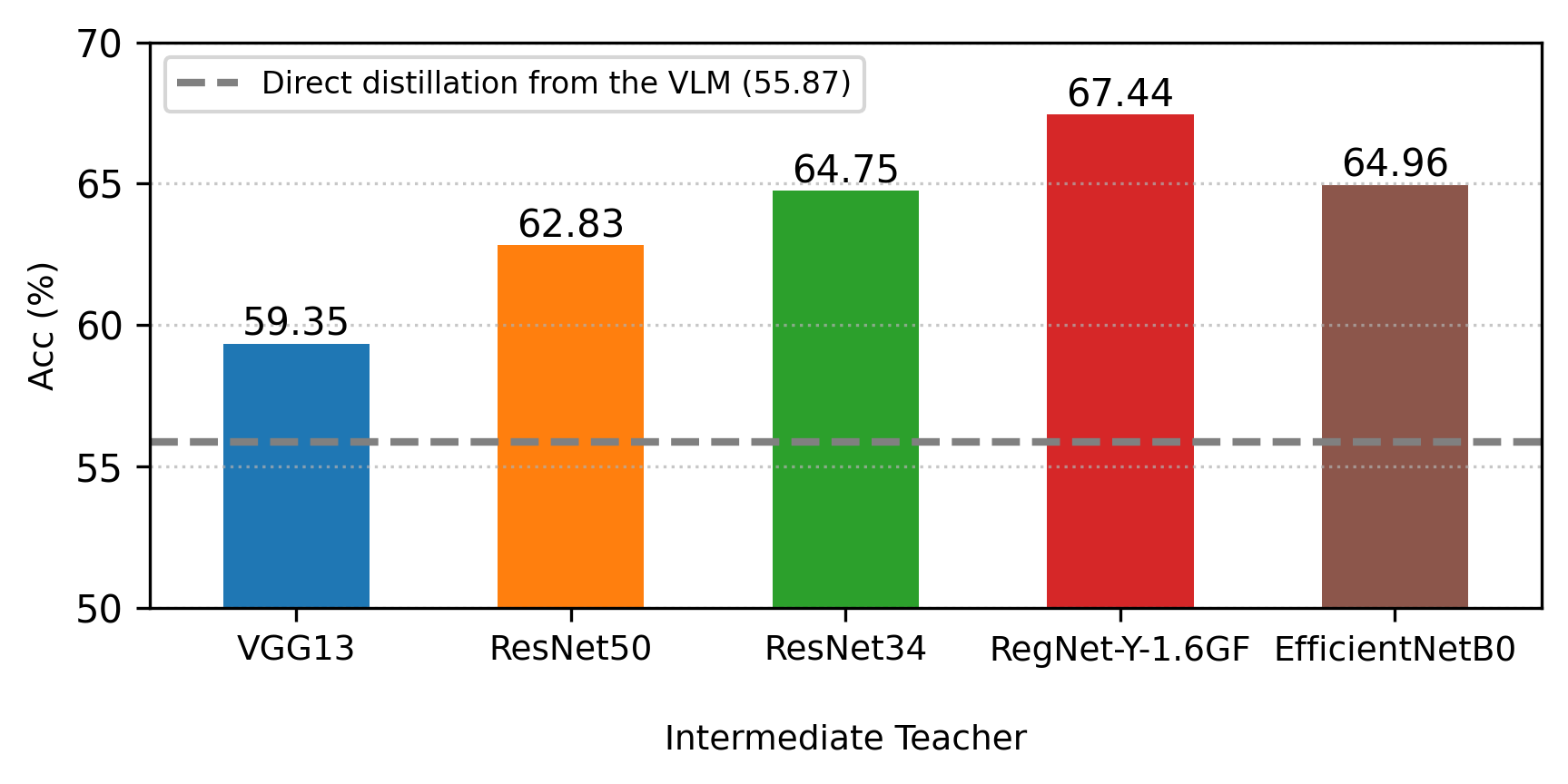}
    \caption{The top-1 accuracy of ResNet18 on FGVC-Aircraft using different intermediate teachers for knowledge distillation. The dashed line represents the baseline result (55.87\%) obtained by direct distillation from the VLM.}
    \label{fig:fig6}
\end{figure*}

\begin{figure*}[!t]
    \centering
    \includegraphics[width=0.75\textwidth]{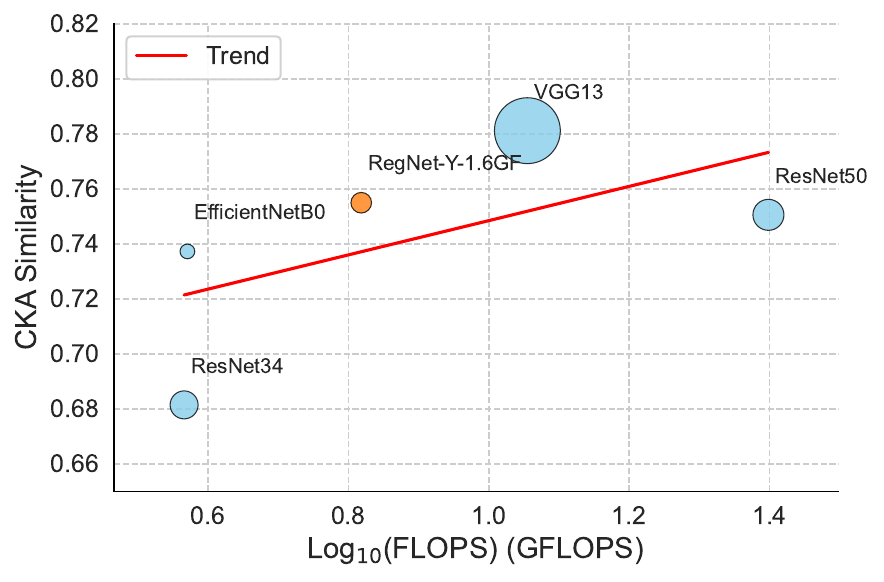}
    \caption{Comparison of CKA feature similarity scores between different intermediate teachers and the VLM. Larger circles denote a greater number of parameters for the corresponding model. Among all candidates, RegNet strikes the optimal balance between model size and architectural compatibility.}
    \label{fig:fig7}
\end{figure*}

Through extensive ablation studies, it can be observed that RegNet-Y-1-6GF consistently achieves the best student performance, as shown in \cref{fig:fig6}. This can be attributed to a favorable match in representation granularity and inductive bias between intermediate teacher and strong teacher: stage-wise width design of RegNet produces feature maps whose scale and channel granularity facilitate feature-level alignment with the VLM, while its modern convolutional blocks offer training stability and efficiency. Interestingly, the CKA analysis\cite{kornblith2019similarity}, as shown in \cref{fig:fig7}, reveals that the intermediate features of VGG-13 exhibit a slightly higher similarity to VLM representations compared to those of RegNet. However, this result may be influenced by the dimensionality of its output features (4096), as high-dimensional features may yield excessively high similarity values in such metrics\cite{murphy2024correcting}. In contrast, RegNet achieves the second-highest CKA value with a much lower feature dimension (888), while also demonstrating better class discriminability. This suggests that its representations strike a more favorable balance between compactness, efficiency, and alignment quality with VLM representations. Unlike prior works that usually use a network larger than the student as a teacher assistant, our DAIT adopts RegNet—even smaller than the student model—as the intermediate teacher. This observation suggests that the selection of an appropriate intermediate teacher may depend more on architectural compatibility rather than solely on the number of parameters.

\subsubsection{The Setting of Hyperparameter $\lambda$.}\label{hyper}
\begin{figure*}[!t]
\centering
\subfloat[]{
  \begin{minipage}[b]{0.46\linewidth}
    \centering 
    \includegraphics[width=1\linewidth]{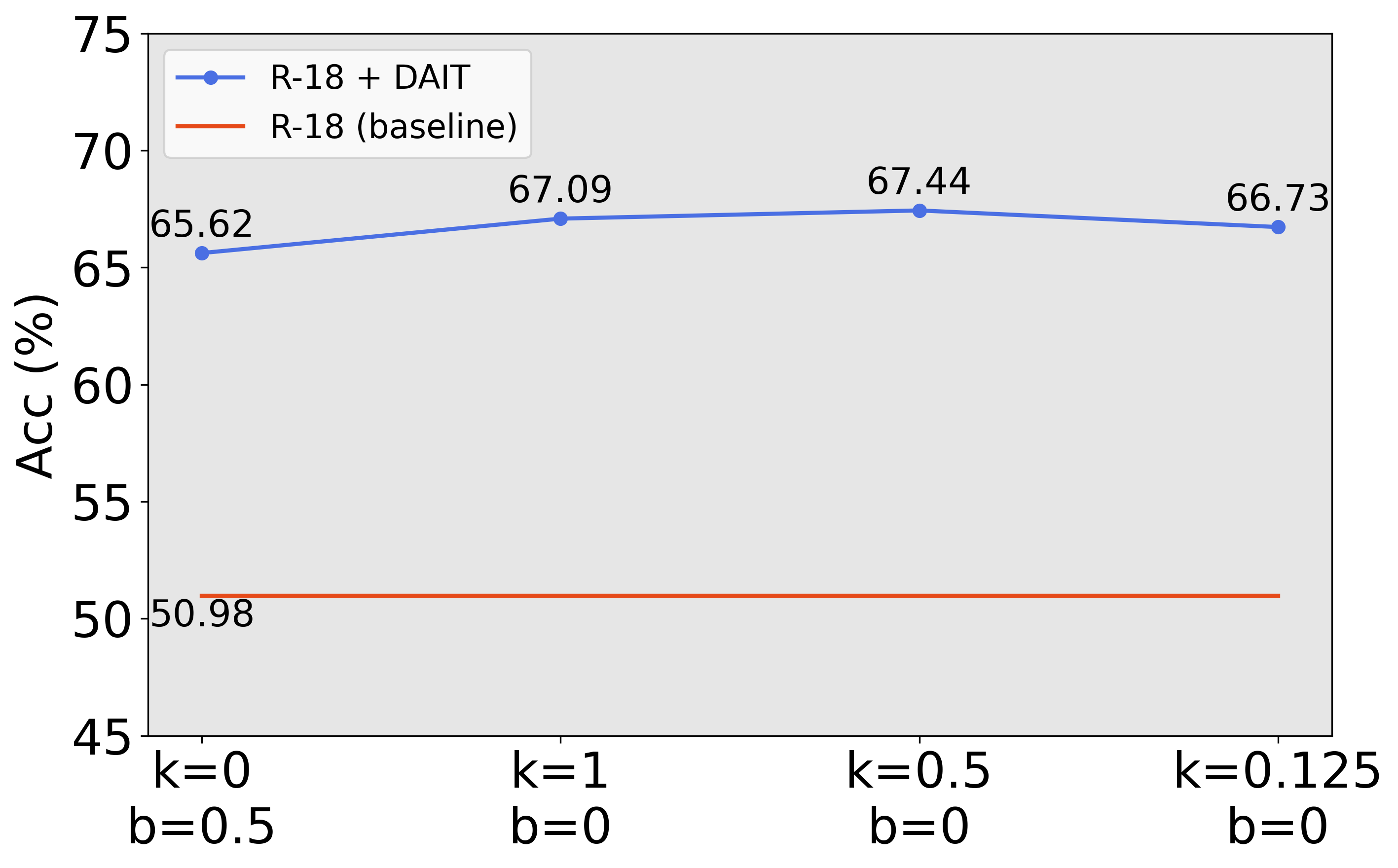}
  \end{minipage}
  \label{fig6a}
}
\subfloat[]{
  \begin{minipage}[b]{0.46\linewidth}
    \centering
    \includegraphics[width=1\linewidth]{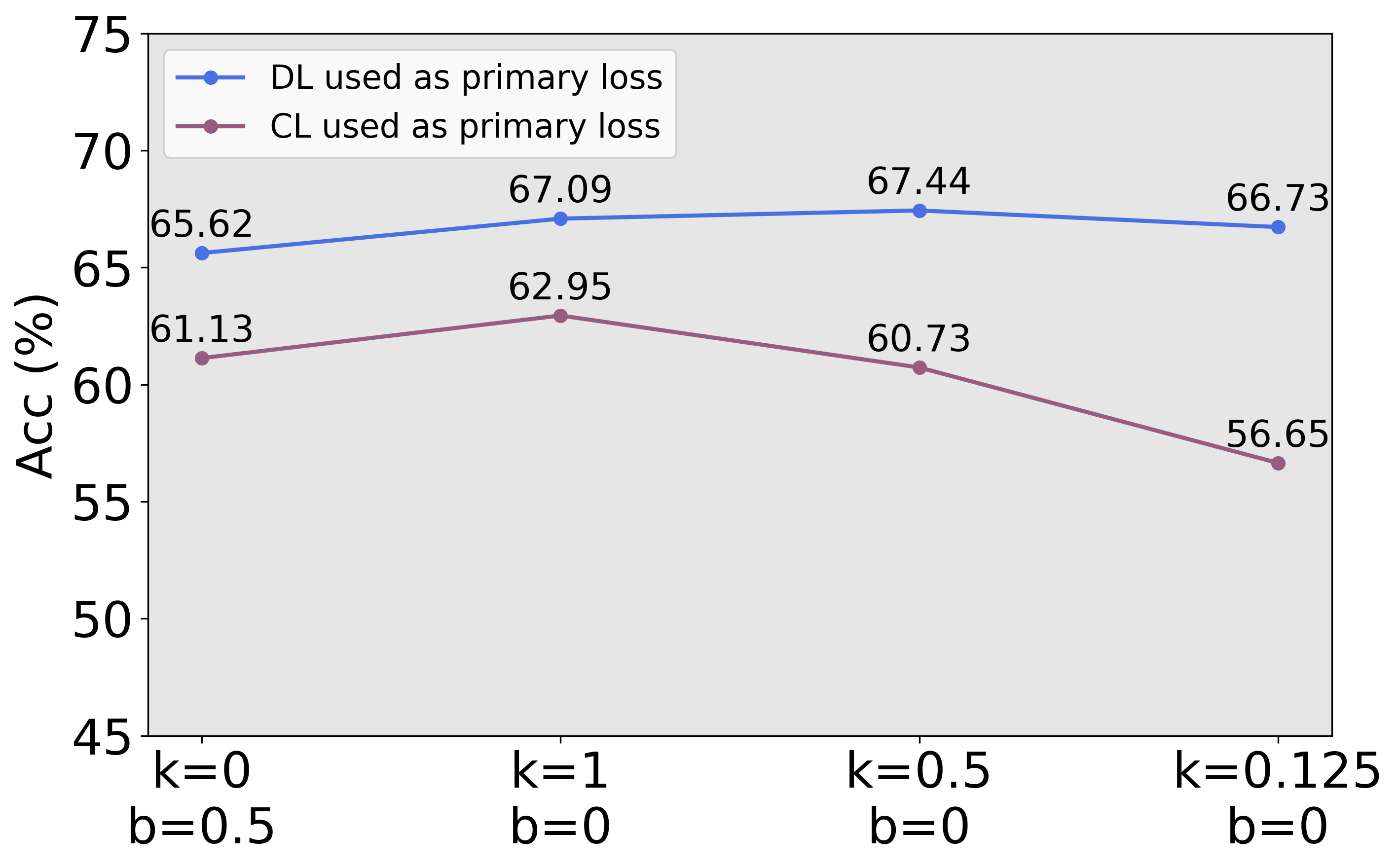}
  \end{minipage}
  \label{fig6b}
}
\caption{The result of different training strategies on the FGVC-Aircraft dataset using ResNet-18. (a) Classification accuracy comparison with  distillation loss adopted as primary loss under different hyperparameter configurations. (b) Classification accuracy comparison with distillation loss (DL) as primary loss against classification loss (CL) as primary loss under different hyperparameter configurations. Here, $k=0$ denotes the loss with fixed weights, and $\lambda$ is directly applied to the secondary loss.}
\label{fig:CLS_DIS}
\end{figure*}

During training, we introduce a hyperparameter $\lambda$ to dynamically balance the distillation loss and the classification loss, allowing each term to play an appropriate role at different training stages. In this section, we analyze in detail the impact of different values of $k$ and $b$ in \cref{eq11} on the performance, as shown in \cref{fig6a}.

On the FGVC-Aircraft dataset, ResNet-18 is adopted as a baseline model to compare different training strategies that treat the distillation loss or the  classification loss as the primary loss. The experimental results are shown in \cref{fig6b}. The results demonstrate that taking the distillation loss as the primary loss in the early training stage effectively guides the model to learn the knowledge from the teacher network, while gradually increasing the contribution of the classification loss in later stages enables more efficient task-adaptive knowledge transfer, leading to the best recognition performance.

\subsubsection{Learning Efficiency in Data Shortages.}The quantity of
training data is an important factor that significantly influences model performance. However, practical applications are likely to be vulnerable to the data scarcity which may compromise model performance. To address this, our research presents some evaluations of the performance of the proposed method with limited training data, as shown in \cref{tab:ratio_improvement}. Specifically, we have assessed the impact of training ResNet-18 with only 30\%, 50\%, and complete training data, implementing DAIT to evaluate its robustness in data-deficient environments. The results indicate that our proposed DAIT substantially enhances performance on FGVC tasks even when the available training data is limited. Moreover, we observe that DAIT yields larger performance gains as the amount of training data decreases.

\begin{table}[!t]
  \centering
  \caption{Comparative results of ResNet-18 between DAIT and the suboptimal method across multiple datasets using 30\%, 50\%, and 100\% of the training data.}
  \label{tab:ratio_improvement}
  \begin{tabular}{ccccc}
    \toprule
    \multirow{2}{*}{Ratio} & \multirow{2}{*}{Method} & \multicolumn{3}{c}{Dataset} \\
    & & CUB-200 & Aircraft & Sf Cars \\
    \midrule
    \multirow{3}{*}{30\%} & VL2Lite & 50.32 & 33.72 & 43.36 \\
    & \textbf{DAIT} & \textbf{66.33} & \textbf{44.32} & \textbf{72.64} \\
    & Diff. & \textcolor{red}{+16.01} & \textcolor{red}{+10.60} & \textcolor{red}{+29.28} \\
    \midrule
    \multirow{3}{*}{50\%} & VL2Lite & 60.87 & 45.18 & 60.41 \\
    & \textbf{DAIT} & \textbf{74.39} & \textbf{54.87} & \textbf{81.85} \\
    & Diff. & \textcolor{red}{+13.52} & \textcolor{red}{+9.69} & \textcolor{red}{+21.44} \\
    \midrule
    \multirow{3}{*}{100\%} & VL2Lite & 71.38 & 55.87 & 77.09 \\
    & \textbf{DAIT} & \textbf{79.77} & \textbf{67.44} & \textbf{88.96} \\
    & Diff. & \textcolor{red}{+8.39} & \textcolor{red}{+11.57} & \textcolor{red}{+11.87} \\
    \bottomrule
  \end{tabular}
\end{table}

\section{Conclusion}
In this paper, we propose DAIT, a distillation framework for fine-grained visual classification, enabling efficient and task-adaptive knowledge transfer from VLMs to lightweight student networks. By introducing a trainable intermediate teacher as a task-adaptation layer, DAIT reorganizes and refines the multimodal representations of VLMs into more compact and discriminative knowledge tailored to the target task. This distillation strategy effectively alleviates representation mismatch, enhances training stability, and improves both learning efficiency and generalization, while avoiding costly fine-tuning of large VLMs. Extensive experiments demonstrate the effectiveness of DAIT on fine-grained visual classification tasks. These results highlight the potential of adaptive intermediate teachers as a practical and scalable solution for deploying powerful VLM knowledge into compact models.

\bibliographystyle{splncs04}

\bibliography{main}

\par\vfill\par
 
\clearpage  


%
%

\end{document}